\documentclass{article}
\usepackage{spconf,amsmath,graphicx}
\usepackage{bm}
\usepackage{amssymb}
\usepackage{color}
\usepackage{multirow}
\usepackage{subfigure}
\usepackage[subfigure]{tocloft}
\usepackage{booktabs}
\usepackage{diagbox}
\usepackage[numbers]{natbib}
\usepackage{enumitem}
\usepackage{tabularx}
\usepackage{arydshln}
\usepackage{caption}

\setenumerate[1]{itemsep=0pt,partopsep=0pt,parsep=\parskip,topsep=0pt}
\setitemize[1]{itemsep=0pt,partopsep=0pt,parsep=\parskip,topsep=0pt}
\setdescription{itemsep=0pt,partopsep=0pt,parsep=\parskip,topsep=0pt}
\title{Attentive Semantic Exploring for Manipulated Face Detection}
\name{Zehao Chen, Hua Yang\sthanks{Corresponding author.}}
\address{Institution of Image Communication and Network Engineering,\\Shanghai Jiao Tong University, Shanghai, China \\
Shanghai Key Lab of Degital Media Processing and Transmission, Shanghai, China\\
}
\begin{document}
\topmargin=0mm

\ninept

\makeatother

\maketitle
%

\begin{abstract}
Face manipulation methods develop rapidly in recent years, whose potential risk to society accounts for the emerging of researches on detection methods. However, due to the diversity of manipulation methods and the high quality of fake images, detection methods suffer from a lack of generalization ability. To solve the problem, we find that segmenting images into semantic fragments could be effective, as discriminative defects and distortions are closely related to such fragments. Besides, to highlight discriminative regions in fragments and to measure contribution to the final prediction of each fragment is efficient for the improvement of generalization ability. Therefore, we propose a novel manipulated face detection method based on Multilevel Facial Semantic Segmentation and Cascade Attention Mechanism. To evaluate our method, we reconstruct two datasets: GGFI and FFMI, and also collect two open-source datasets. Experiments on four datasets verify the advantages of our approach against other state-of-the-arts, especially its generalization ability.

\end{abstract}
\section{Introduction}
\label{sec:intro}

With the rapid development of deep learning, face image and video manipulation methods have made great progress. State-of-the-art manipulation methods can generate fake face images with such high quality that it is difficult for humans to distinguish. However, face manipulation methods may be maliciously exploited by lawbreakers, causing severe security problems to the society, such as the wide spread of fake news and the rising risk of privacy safety. 

At present, the mainstream face manipulation methods can be divided into three categories according to their functions: face generation, facial features manipulation, and face swap. Face generation is generating face images directly using Generative Adversarial Networks (GANs)\cite{NIPS2014_5423} or Variational AutoEncoders (VAEs) \cite{kingma2013auto}. State-of-the-art GANs, such as PGGAN \cite{karras2017progressive}, SGAN, \cite{Karras_2019_CVPR,Karras_2020_CVPR} and MSGGAN\cite{Karnewar_2020_CVPR} can generate high-quality face images with the resolution of $1024 \times 1024$. Facial features manipulation is to change facial attributes on real face images, like hair color, hairstyle, gender, expression, and others. StarGAN \cite{Choi_2018_CVPR} and StarGANv2 \cite{Choi_2020_CVPR} can change facial features automatically after setting parameters, and SC-FEGAN \cite{Jo_2019_ICCV} can achieve this function through drawing masks by users. Face swap can be separated into two varieties: identity swap and expression swap. Identity swap replaces target person's entire face by source's, so the identity of target is changed to source's. DeepFakes \cite{DeepFake} and FaceSwap \cite{FaceSwap} are popular applications that can achieve this function. Expression swap just changes target's expression by source's, but does not change target's identity, which is widely used in video games and movies \cite{thies2015real,thies2016face2face,thies2019deferred}. Scholars also present several open-source face swap datasets \cite{rossler2019faceforensics++,Li_2020_CVPR_celeb}.

In response to the rapid development of face manipulation methods, scholars gradually notice the necessity and importance of detecting manipulated faces and put forward several detection methods. Methods proposed by early studies \cite{zhou2017two,mo2018fake,afchar2018mesonet,10.1145/3335203.3335724} are usually designed for a certain category of manipulated face images. They are not well equipped to detect nowaday diverse types of fake images. Scholars try to tackle diverse types of fake face images with multifarious ideas in recent studies\cite{multi-task,Li_2020_CVPR,Dang_2020_CVPR,nguyen2019capsule,nguyen2019use,chai2020makes,qian2020thinking}. For instance, \cite{multi-task} proposes an auto-encoder-based model to detect manipulated face images. \cite{Dang_2020_CVPR} puts forward an attention-based CNN to locate manipulation regions in fake images. \cite{nguyen2019capsule,nguyen2019use} use dynamic routing algorithm to choose features extracted by several Capsule Networks. \cite{qian2020thinking} mines frequency-aware clues for detection. And \cite{chai2020makes} designs a patch-based classifier with limited receptive fields to visualize which regions of fake images are more easily detectable. However, due to the high quality of fake images and diversity of manipulation methods, state-of-the-art detection methods still suffer from a problem: lack of generalization ability. An excellent detector is desired to perform well not only on different varieties of fake images but also on images generated by unseen manipulation methods. Therefore, we seek to develop a detection method to tackle high-quality fake images and various manipulation methods.

\setlength{\abovecaptionskip}{3pt}
\setlength{\belowcaptionskip}{0pt}

\begin{figure}[t]
  \centering
   \subfigure[]{\includegraphics[width=0.24\linewidth]{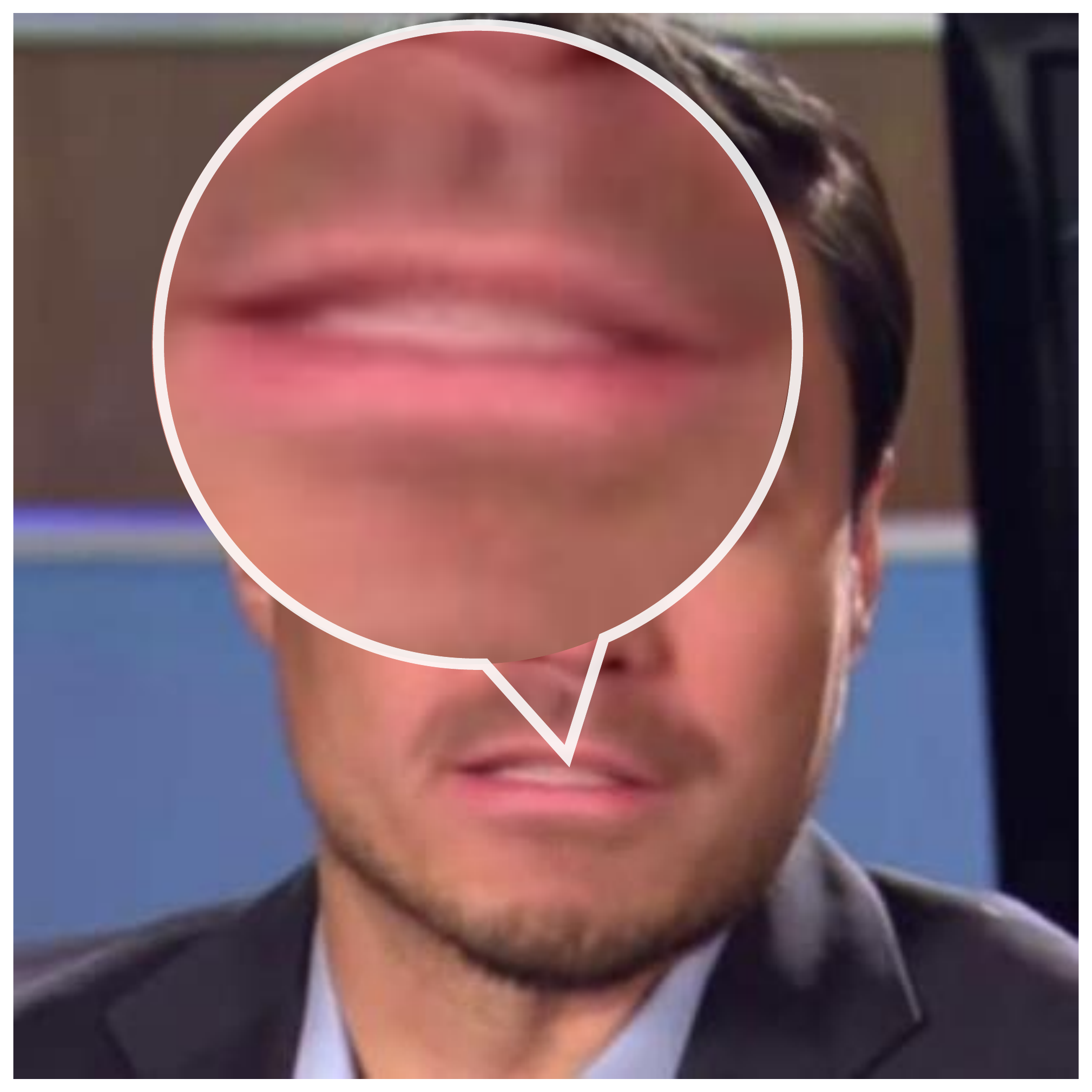}\label{fig:introa}}
   \subfigure[]{\includegraphics[width=0.24\linewidth]{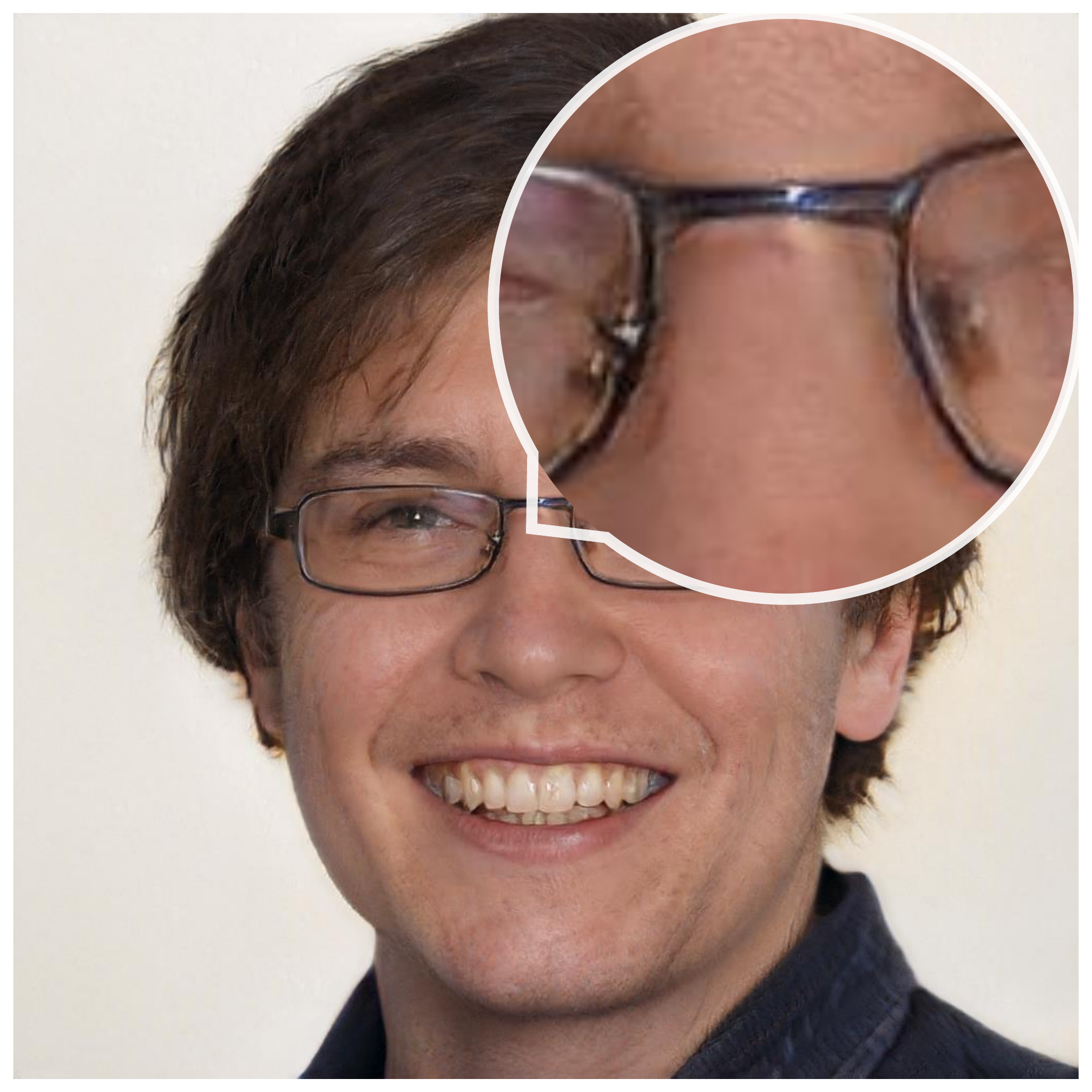}\label{fig:introb}}
   \subfigure[]{\includegraphics[width=0.24\linewidth]{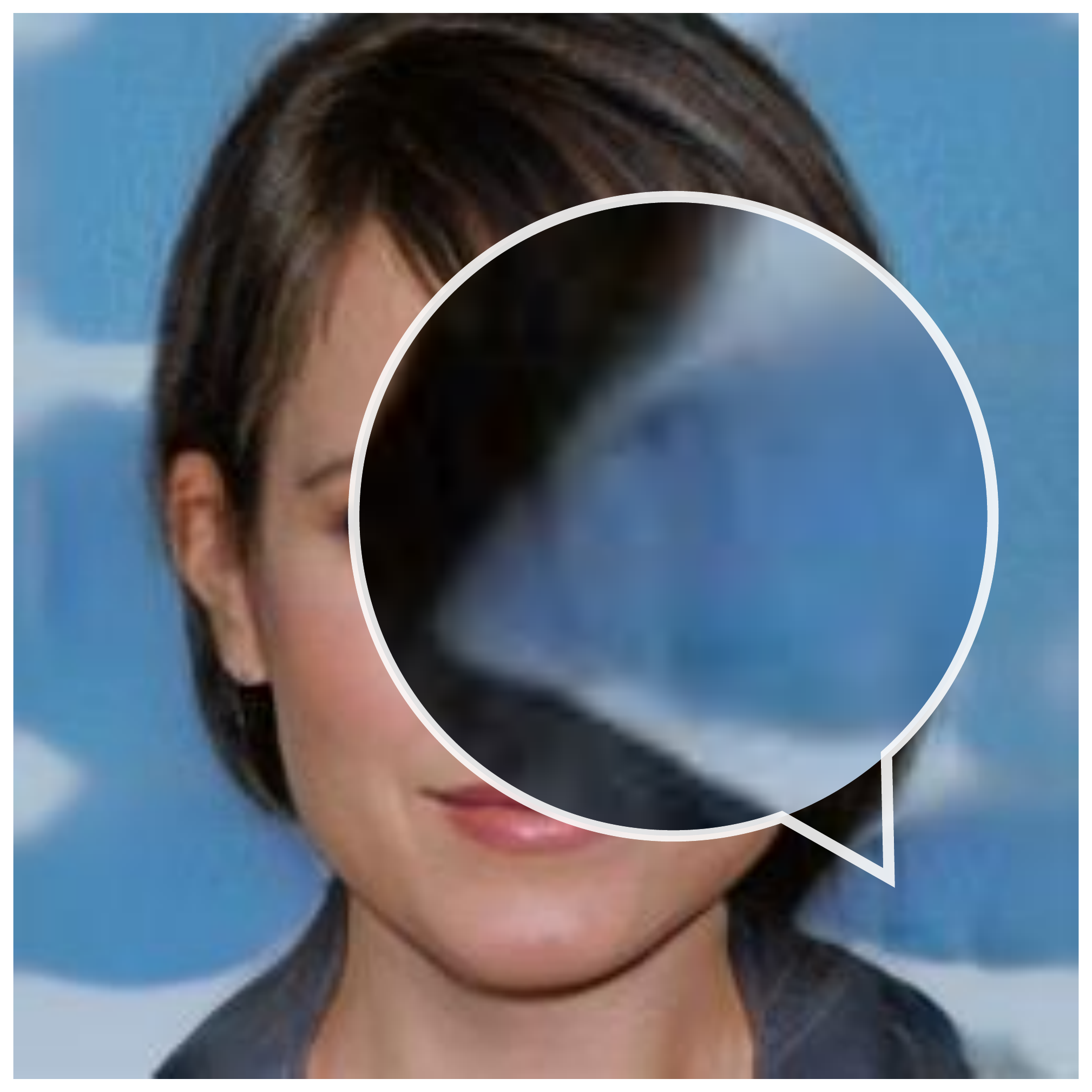}\label{fig:introc}}
   \subfigure[]{\includegraphics[width=0.24\linewidth]{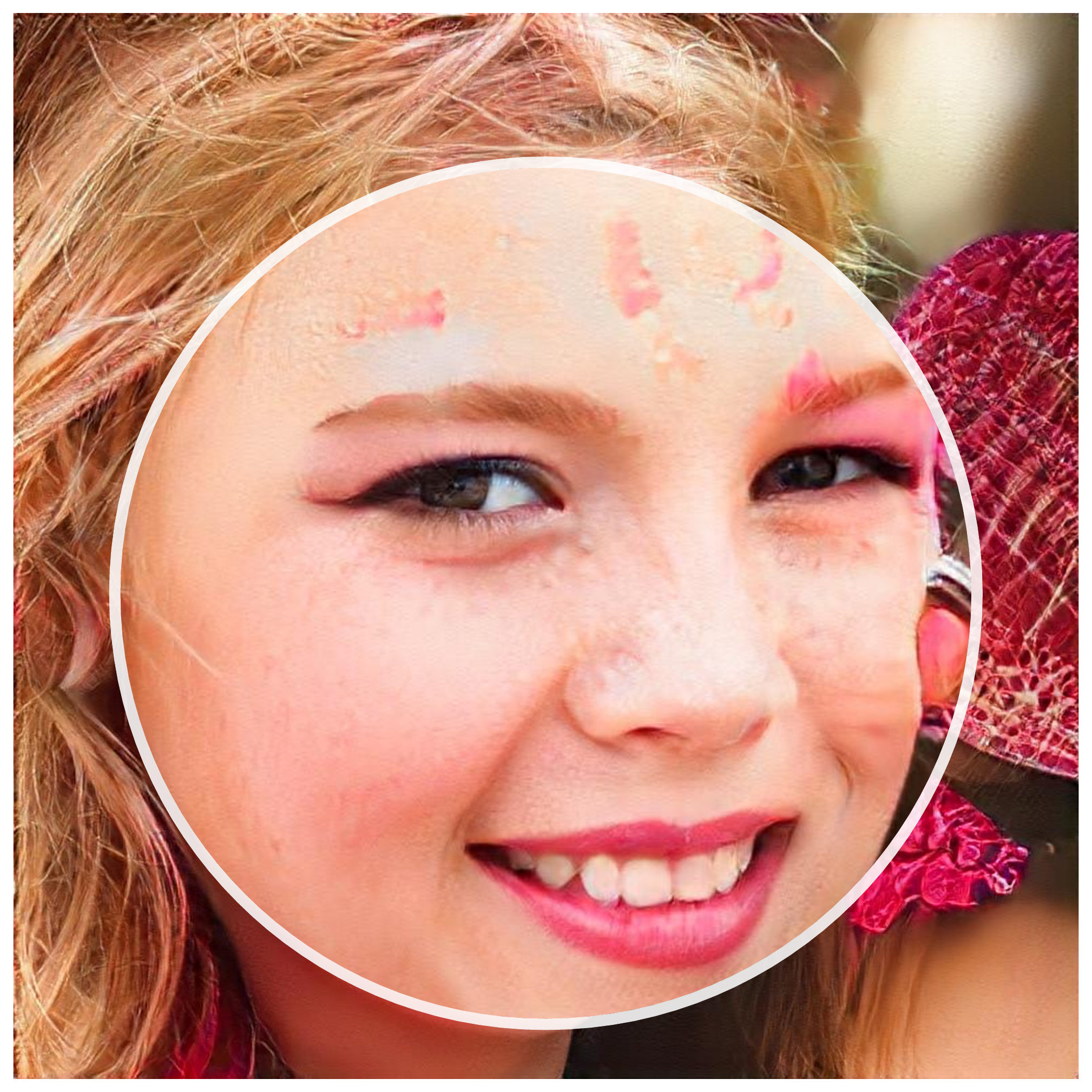}\label{fig:introd}}
      \caption{Defects and distortions in faces images generated by state-of-the-art manipulation methods: (a) defect on teeth, (b) defect on glasses, (c) distortion in background, (d) distortion in face.}
 \label{fig:intro}
\end{figure}

Although fake face images generated by various latest methods are generally of high quality, they still have tiny defects in the details, even the defects difficult to be perceived by human eyes. We find that these defects are the key to distinguishing real and fake images and improving the generalization ability of the detector. Usually, defects are closely related to semantic features, appearing in local regions containing key information in face images, such as eyes, nose, and mouth, as illustrated in Fig.\ref{fig:introa} and \ref{fig:introb}. Thus, such semantic local areas are discriminative that can be segmented into independent fragments for detection. Meanwhile, fake images may be slightly distorted globally, as illustrated in Fig.\ref{fig:introc} and \ref{fig:introd}. Global distortions indicate that using global fragments for detection is also helpful. Therefore, we investigate a Multilevel Facial Semantic Segmentation (MFSS) approach to generate local and global semantic fragments for detection.

The diversity of manipulation methods is also a key factor limiting the generalization ability. Fake images generated by different manipulation methods differ visually and statistically, which makes it difficult for the detector to identify fake images generated by unseen manipulation methods. Aiming to solve this, we present a Local Attention Module (LAM) to highlight generalized discriminative regions in semantic fragments. After using local and global semantic fragments for detection, it's necessary for the detector to have a precise measurement of how important a role each fragment plays. For example, if a fragment of a fake image contains too few fake features to be predicted as fake, it must contribute little to the final prediction. To this end, we design a Semantic Attention Module (SAM). 

Motivated by the above findings, we present a novel manipulated face detection method with Multilevel Facial Semantic Segmentation and Cascade Attention Mechanism. MFSS segments a face image into six semantic fragments to provide local (eyes, nose, and mouth) and global (background, face, and the image itself) semantic regions for detection. Cascade Attention Mechanism consists of Local Attention Modules and Semantic Attention Module. LAMs help the method focus more on generalized discriminative regions and SAM measures how much each semantic fragment contributes to the final prediction. Experiments on four datasets and comparison with state-of-the-arts demonstrate superior generalization ability of our method. Our main contributions are as follows:
\begin{itemize}[leftmargin = 0pt, itemindent = 25pt]
    \item We present a novel manipulated face detector framework based on Multilevel Facial Semantic Segmentation to provide six local and global semantic fragments for detection.
    \item We propose a Cascade Attention Mechanism consisting of LAMs and SAM to respectively highlight generalized discriminative regions and measure each fragment's contribution.
    \item We reconstruct two datasets for method evaluation: GGFI and FFMI. Experiments on the above datasets as well as two public datasets demonstrate our method exceeds state-of-the-arts, especially on the generalization ability.
\end{itemize}

\section{Proposed Method}

\subsection{Overview}

\begin{figure}[t]
  \centering
   \includegraphics[width=0.99\linewidth]{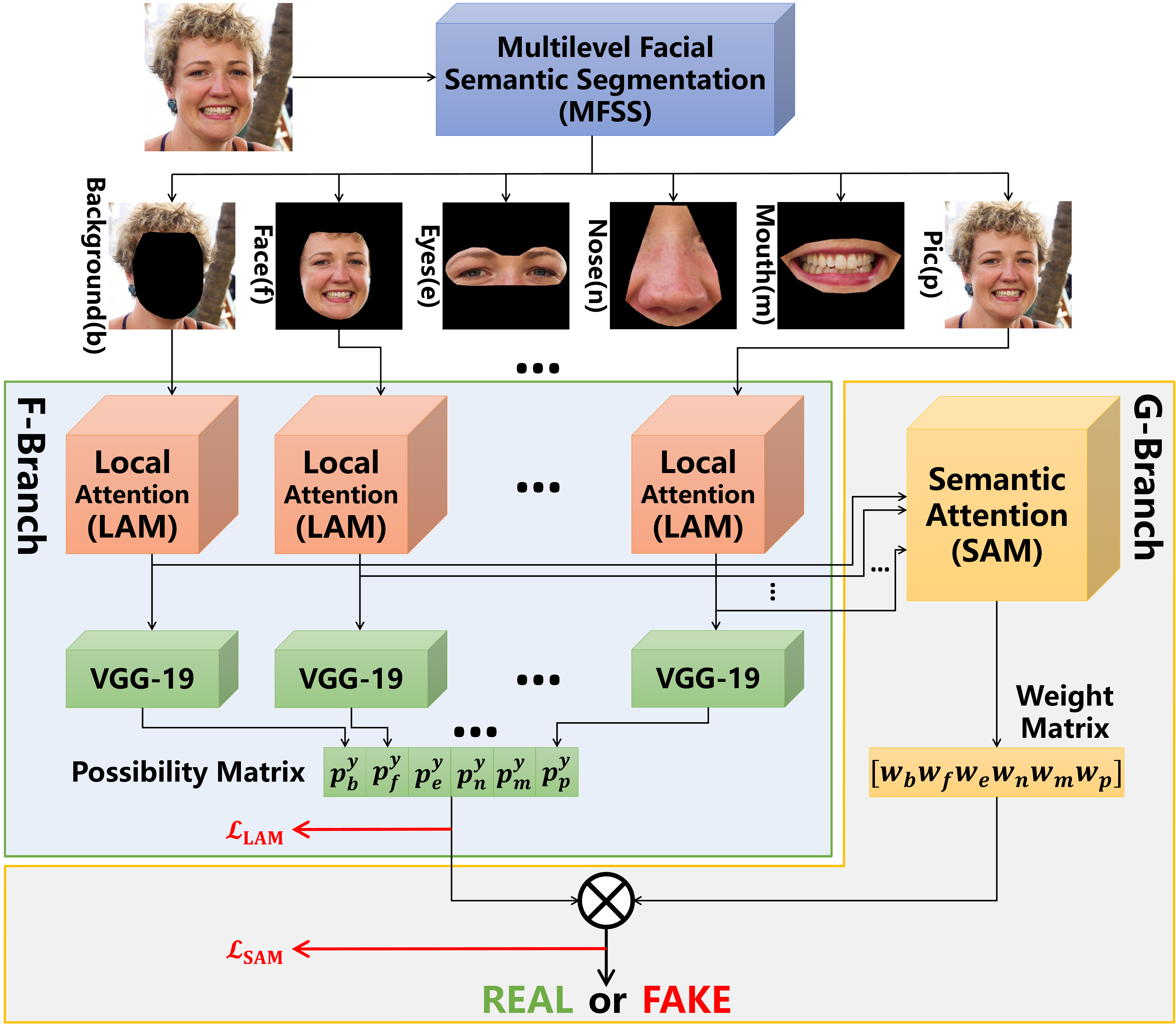}
      \caption{The architecture of proposed method: where $\bigotimes$ denotes matrix multiplication.}
 \label{fig:model}
\end{figure}

As illustrated in Fig.\ref{fig:model}, our proposed method mainly consists of three parts, Multi-level Facial Semantic Segmentation (MFSS), Fragment Branch (F-Branch), and Global Branch (G-Branch). Firstly, MFSS segments a face image into six semantic fragments in two steps, which are background, face, eyes, nose, mouth, and itself, denoting as $b,f,e,n,m,$ and $p$ respectively. The first step of MFSS is to extract 81 facial landmarks by dlib face detector \cite{dlib}. And the second step of MFSS is to group and connect facial landmarks to generate six semantic fragments. Subsequently, F-Branch with Local Attention Modules (LAMs) for fragment-level supervision classifies six fragments respectively. Finally, G-Branch with Semantic Attention Module (SAM) for global supervision generates weight for each fragment and produces the final prediction.

\subsection{Fragment Branch}

F-Branch is used to binarily classify the six fragments of face images, which consists of six sets of Local Attention Modules and backbone networks. The input is six fragments of face images. Each fragment passes through its own set of LAM and backbone network to produce the corresponding possibility vector $\bm{p_{i}^{y}} = [p_{i}^{0},p_{i}^{1}]^ \mathrm{T}$, where $p_{i}^{0},p_{i}^{1}$ denote the output of backbone networks after a softmax operation to normalize the logits (the possibility of fake and real) and $i$ denotes six fragments $(i \in \{b,f,e,n,m,p\})$. The output of F-Branch is a matrix called Possibility Matrix : $\mathbf{P} = [\bm{p_{p}^{y},p_{b}^{y},p_{f}^{y},p_{e}^{y},p_{m}^{y},p_{n}^{y}}] \in \mathbb{R} ^{2 \times 6}$.

\begin{figure}[t]
  \centering
  \subfigure[LAM]{\includegraphics[width=0.484\linewidth]{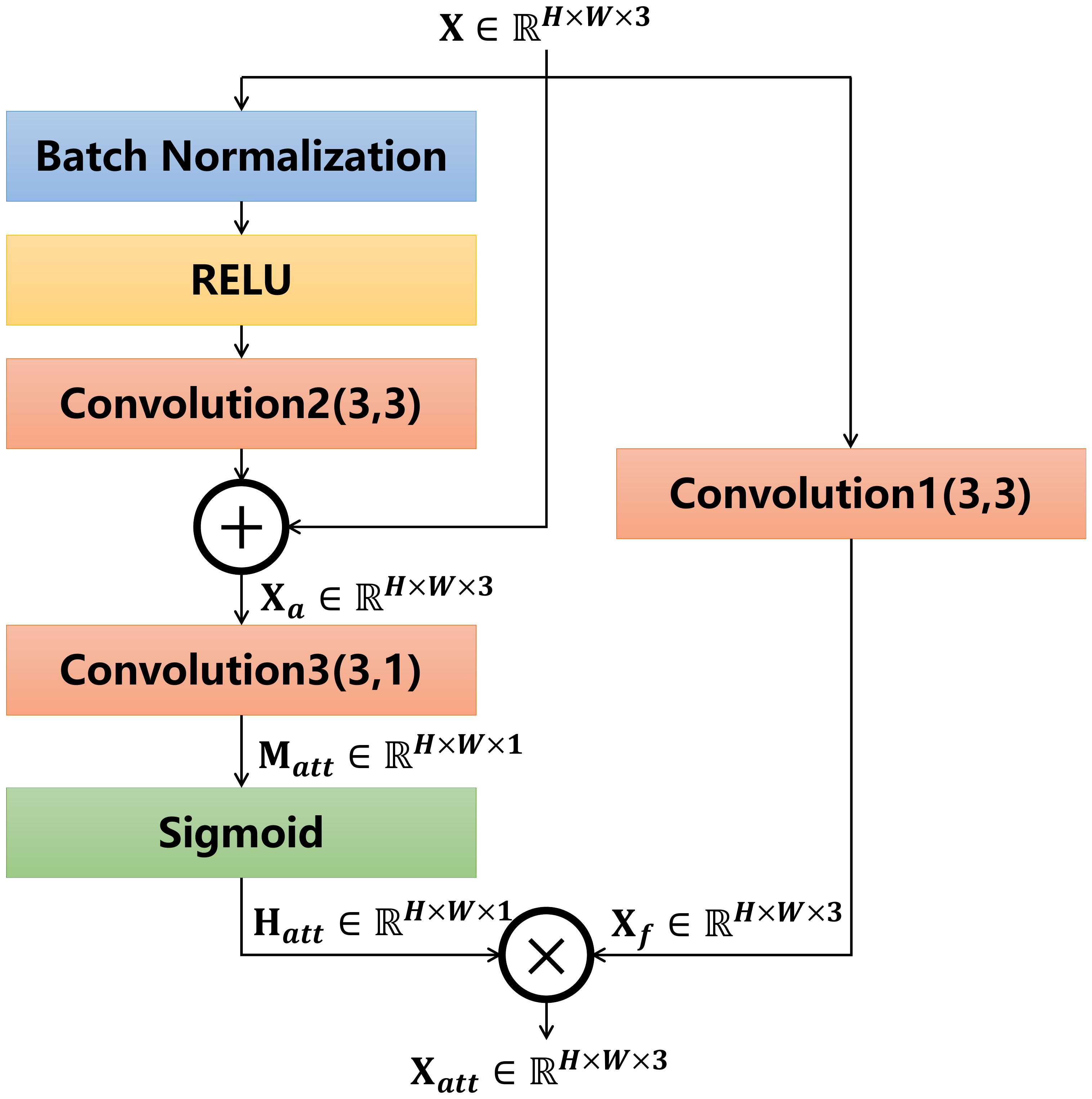}\label{fig:LAM}}
  \subfigure[SAM]{\includegraphics[width=0.496\linewidth]{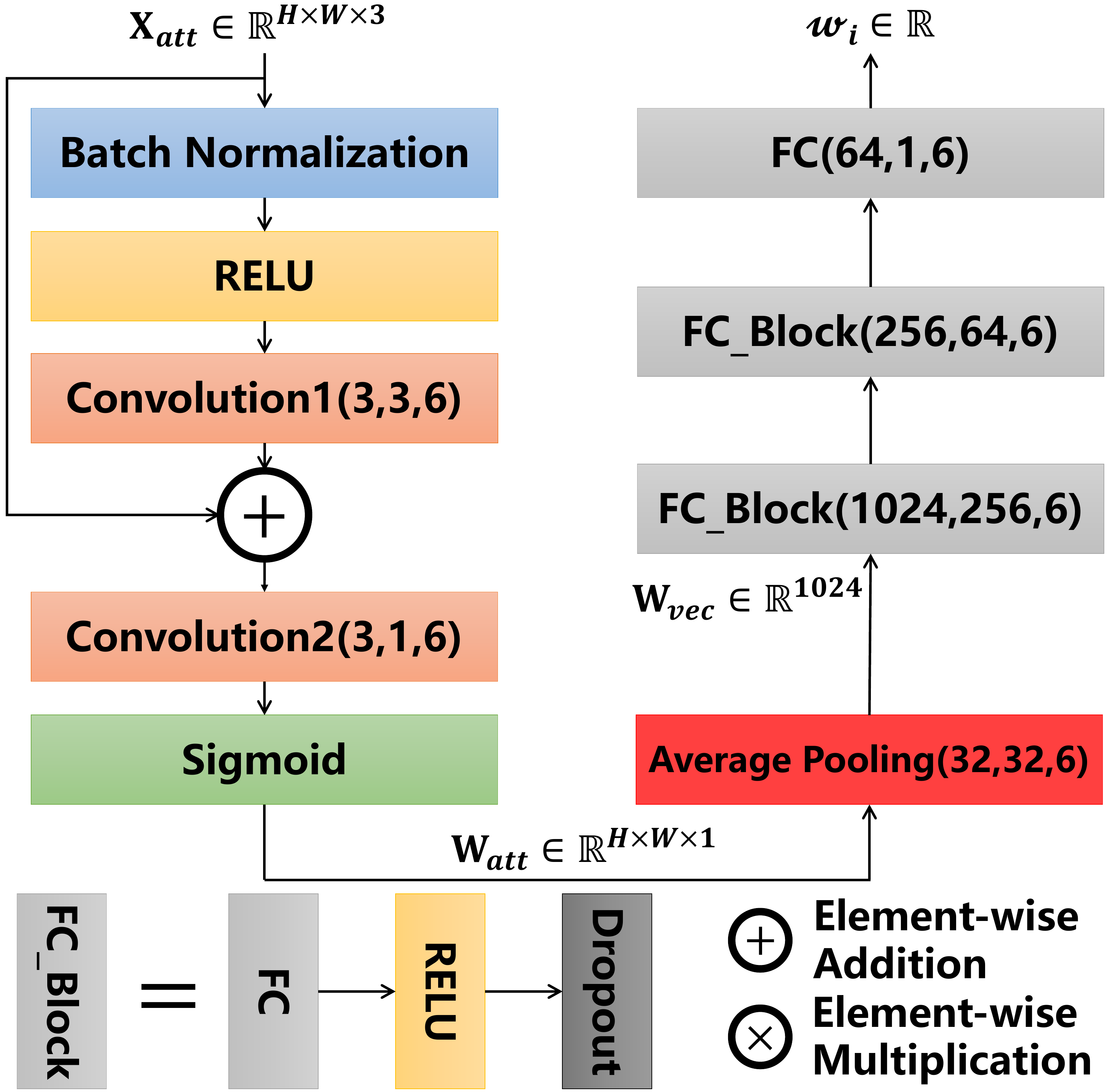}\label{fig:SAM}}
      \caption{The architecture of LAM and SAM: for all convolutional and fully-connected layer, the first parameter is the number of input channels, the second is the number of output channels, and the third is the number of this layer in SAM.}
 \label{fig:attention}
\end{figure}

Given a fragment $\mathbf{X} \in \mathbb{R} ^{H \times W \times 3}$, where H, W, and 3 denote channel, height, and width respectively. We utilize LAM to add attention to $\mathbf{X}$ to generate attentive fragments $\mathbf{X}_{att} \in \mathbb{R} ^{H \times W \times 3}$. As illustrated in Fig.\ref{fig:LAM}, the LAM is comprised of two streams: the feature stream and the attention stream. The feature stream produces an image feature map $\mathbf{X}_{f} \in \mathbb{R} ^{H \times W \times 3}$ via a convolutional operation $Convolution1$: $\mathbf{X} \xrightarrow{Conv1} \mathbf{X}_{f}$, where $Convolution1$ is a convolutional layer with the kernel size of $3\times3$ and the output channels of $3$. As for the attention stream, it firstly produces an attention feature map $\mathbf{X}_{a} \in \mathbb{R} ^{H \times W \times 3}$ via a residual block, inspired by residual used in CNNs \cite{He_2016_CVPR}: $\mathbf{X}_{a} = \mathbf{X} + \mathcal{R}(\mathbf{X})$, where $\mathcal{R}$ denotes $\mathbf{X}\xrightarrow{bn}\xrightarrow{relu}\xrightarrow{Conv2}\mathbf{X}^{'}$, and $Convolution2$ has the same parameters as $Convolution1$. Secondly, attention map $\mathbf{M}_{att} \in \mathbb{R} ^{H \times W \times 1}$ is generated via a convolutional operation $Convolution3$: $\mathbf{X}_{a}\xrightarrow{Conv3}\mathbf{M}_{att}$, where $Convolution3$ is a convolutional layer with the kernel size of $3\times3$ and the output channel of $1$. The third step is to acquire an attention heatmap $\mathbf{H}_{att} \in \mathbb{R} ^{H \times W \times 1}$ by applying the sigmoid function to $\mathbf{M}_{att}$: $\mathbf{H}_{att} = Sigmoid(\mathbf{M}_{att})$. Thus, $\mathbf{M}_{att}$ is normalized from 0 to 1. The higher attention a pixel in $\mathbf{X}$ has, the closer to 1 its corresponding value in $\mathbf{M}_{att}$ is. Finally, the output attentive fragments $\mathbf{X}_{att}$ is generated by doing element-wise multiplication between image feature map $\mathbf{X}_{f}$ and attention heatmap $\mathbf{H}_{att}$: $\mathbf{X}_{att} = \mathbf{X}_{f} \odot \mathbf{H}_{att}$. In summary, LAM highlights discriminative regions in $\mathbf{X}$. 

After passing through LAM, $\mathbf{X}_{att}$ is to be classified by a backbone network. We utilize VGG-19 \cite{simonyan2014very} pre-trained on ImageNet \cite{5206848} as our backbone networks. Each fragment has a prediction generated by corresponding set of LAM and VGG-19: 
\begin{equation}
    \widehat{y_{i}} = arg\max_{y}(p_{i}^{0},p_{i}^{1}).
\end{equation}
When training, each set of LAM and VGG-19 is optimized together by a cross entropy $\mathcal{L}_{\mathbf{LAM}}$:
\setlength\abovedisplayskip{2pt}
\setlength\belowdisplayskip{0pt}
\begin{equation}
    \mathcal{L}_{\mathbf{LAM}}(\widehat{y_{i}}) = \sum_{y}ylog(p_{i}^{y}),
\end{equation}
where $y$ denotes the label of fake or real for binary classification, $y \in \{0,1\}$.

\subsection{Global Branch}

The key module in G-Branch is the Semantic Attention Module. The input of SAM is six attentive fragments of a face image, which are the output of six LAMs in F-Branch. Suppose an attentive fragment $\mathbf{X}_{att}$. It firstly passes through the same structure as the attention stream in LAM to generate a weight attention map $\mathbf{W}_{att} \in \mathbb{R} ^{H \times W \times 1}$: $\mathbf{X}_{att} \xrightarrow{Att\ Stream} \mathbf{W}_{att}$. Subsequently, it produces a weight vector $\mathbf{W}_{vec} \in \mathbb{R}^{1024}$ via an average pooling layer with the output size of $32 \times 32$: $\mathbf{W}_{att} \xrightarrow{Ave\ Pool} \mathbf{W}_{vec}$. Finally, we utilize a three-layer multi-layer perception to generate the final weight $w_{i}$ of $\mathbf{X}_{att}$: $\mathbf{W}_{vec} \xrightarrow{MLP} w_{i}$. SAM comprises six same networks as Fig.\ref{fig:SAM} illustrated for six attentive fragments and produces corresponding weight $w_{i}(i \in \{b,f,e,n,m,p\})$ as the output called Weight Matrix: $\textbf{W} = [w_{p},w_{b},w_{f},w_{e},w_{m},w_{n}] \in \mathbb{R} ^{1 \times 6}$. The final prediction $\widehat{y} \in \{0,1\}$ is generated by multiplication of Possibility Matrix and Weight Matrix:
\begin{equation}
    \widehat{y} = arg\max_{y}(\textbf{PW}^ \mathrm{T}).
\end{equation}
To optimize G-Branch, we apply a cross-entropy $\mathcal{L}_{\mathbf{SAM}}$ to the final prediction:
\begin{equation}
    \mathcal{L}_{\mathbf{SAM}}(\widehat{y}) = \sum_{y}\Big(y\log(\sum_{i}p_{i}^{y}w_{i})\Big), i\in\{b,f,e,n,m,p\},
\end{equation}
where $p_{i}^{y}$ is the output of F-Branch after a softmax operation to normalize the logits, $y$ denotes the label of fake or real for binary classification, $y \in \{0,1\}$.

\section{Experiments}

\subsection{Datasets and Implementation Details}
\label{sec:Dataset}

In order to evaluate the performance of our method in the face of diverse of manipulated face images, we reconstruct two datasets, which are GAN-Generated Face Images (GGFI) and Facial Features Manipulation Images (FFMI). GGFI is a face generation dataset reconstructed by us, which contains real face images from CelebA-HQ\cite{karras2017progressive} and FFHQ\cite{Karras_2019_CVPR}, as well as fake images generated by four state-of-the-art GANs: PGGAN\cite{karras2017progressive}, SGAN\cite{Karras_2019_CVPR}, SGAN2\cite{Karras_2020_CVPR}, and MSGGAN\cite{Karnewar_2020_CVPR}. All of GANs are pre-trained on CelebA-HQ and FFHQ. FFMI is a facial features manipulation dataset reconstructed by us, which contains real face images from CelebA\cite{Liu_2015_ICCV} and CelebA-HQ\cite{karras2017progressive}, and manipulated images generated by two state-of-the-art methods to modify facial features: StarGAN\cite{Choi_2018_CVPR} pre-trained on CelebA and StarGANv2\cite{Choi_2020_CVPR} pre-trained on CelebA-HQ. Besides, we collect two open-source datasets FaceForensics++ (FF++)\cite{rossler2019faceforensics++} and Celeb-DF (C-DF)\cite{Li_2020_CVPR_celeb}. They are both public face swap datasets in the form of videos. We extract key frames from videos, and use dlib face detector\cite{dlib} to get the images of face region. The specified composition of four datasets' training sets is summarized in Table.\ref{tab:datasets}. Their respective test sets have the same composition as training sets and the size of validation sets is a tenth of training sets. 

\begin{table}[t]
\setlength\tabcolsep{0.1pt}

\newcommand{\tabincell}[2]{\begin{tabular}{@{}#1@{}}#2\end{tabular}}
  \caption{The composition of four datasets' training sets: GGFI and FFMI are collected and reconstructed by us. FF++ and C-DF are open-source datasets.}
  \label{tab:datasets}
  \centering
  \ninept
  \begin{tabular}{c|c|c} 
  \hline
     \multirow{2}{*}{Dataset} & \multicolumn{2}{c}{The number of images} \\
     \cline{2-3}
     & Real & Manipulated\\
     \hline
     GGFI & \tabincell{c}{CelebA-HQ\cite{karras2017progressive}:10000\\FFHQ\cite{Karras_2019_CVPR}:10000} & \tabincell{c}{PGGAN\cite{karras2017progressive}:5000,SGAN2\cite{Karras_2020_CVPR}:5000\\SGAN\cite{Karras_2019_CVPR}:5000,MSGGAN\cite{Karnewar_2020_CVPR}:5000}\\
     \hline
     FFMI & \tabincell{c}{CelebA-HQ\cite{karras2017progressive}:10000\\CelebA\cite{Liu_2015_ICCV}:10000} & \tabincell{c}{StarGAN\cite{Choi_2018_CVPR}:10000\\StarGANv2\cite{Choi_2020_CVPR}:10000}\\
     \hline
     FF++\cite{rossler2019faceforensics++} & 20000 & \tabincell{c}{FS\cite{FaceSwap}:5000, DF\cite{DeepFake}:5000\\F2F\cite{thies2016face2face}
     :5000, NT\cite{thies2019deferred}:5000}\\
     \hline
     C-DF\cite{Li_2020_CVPR_celeb} & 20000 & 20000\\
    \hline
  \end{tabular}
\end{table}

The training of our proposed method is in two steps. The first step is to train F-Branch, where six sets of LAMs and VGG-19s are optimized respectively with the corresponding input fragments by the loss function $\mathcal{L}_{\mathbf{LAM}}$. The second step training is for G-Branch, which is optimized by the loss function $\mathcal{L}_{\mathbf{SAM}}$. For both two-step training, we use SGD as the optimizer with a momentum rate of 0.9 and an initial learning rate of $10^{-3}$. The learning rate decay is set as a factor of 0.1 for every 5 epochs during 15 epochs in total. We choose the model with the best performance on the validation set.

\subsection{Comparison with State-of-the-art Methods}
\begin{figure}[t]
  \centering
  \includegraphics[width=0.85\linewidth]{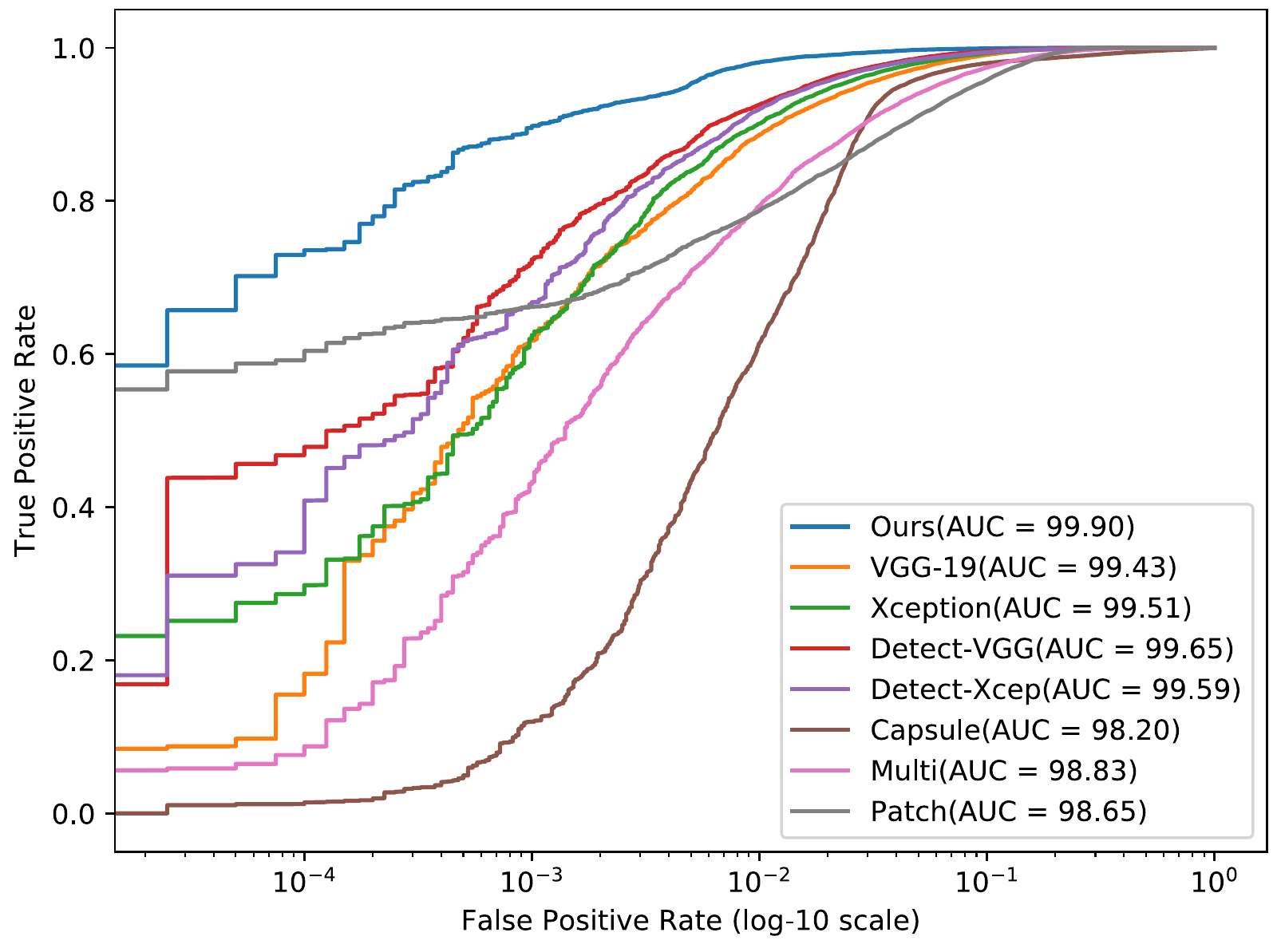}
      \caption{Manipulated face images detection ROC curves of Ours and state-of-the-arts on Merge dataset: our method has the highest AUC.}
 \label{fig:AUC}
\end{figure}

Firstly, we verify the ability of our proposed method to detect diverse types of manipulated face images. Experiments are respectively on the four datasets mentioned in Section \ref{sec:Dataset} and a Merge dataset containing all of them. Merge dataset contains 11 types of manipulation methods, which can effectively evaluate the ability to detect diverse types of fake images. Table.\ref{tab:Test on Seen Data} and show a comparison of our model with other state-of-the-art methods. We can find that our method has the highest accuracy on each dataset. Especially, our approach outperforms others on Merge dataset with over 1\% exceeding on accuracy, and we also present ROC curves on Merge dataset shown in Fig.\ref{fig:AUC}. The results reveal that our method has the best ability to detect diverse types of fake face images against other state-of-the-arts.

\begin{table}[h]
\setlength\tabcolsep{3pt}
  \caption{Experiment results and comparison with other methods.}
  \label{tab:Test on Seen Data}
  \centering
  \ninept
  \begin{tabular}{l|c|c|c|c|c} 
  \hline
    Method & GGFI & FFMI & FF++ & C-DF  & Merge  \\
    \hline
     VGG-19\cite{simonyan2014very} & 99.48 &  99.50 & 99.69  & 94.37  &97.16 \\
     Xception\cite{Chollet_2017_CVPR} & 98.19 & 99.47 & 99.16 & 93.92 & 97.12 \\
     Detect-VGG-19\cite{Dang_2020_CVPR} & 99.73 & 99.83 & 99.79 & 94.79 & 97.24\\
     Detect-Xception\cite{Dang_2020_CVPR} & 99.63 & 99.76 & 99.60 & 94.09  &  96.65\\
     Capsule\cite{nguyen2019use} & 96.53& 99.88 & 98.17 & 94.14 & 95.48\\
     Multi\cite{multi-task} & 99.77&99.82 & 99.80 & 92.15& 95.24\\
     Patch\cite{chai2020makes} & 99.86 & 99.87 & 99.77 & 84.88 & 92.69 \\
     \textbf{Ours}    & \textbf{99.94} & \textbf{99.98} &  \textbf{99.95} & \textbf{95.75}  & \textbf{98.65}\\
     \hline
  \end{tabular}
\end{table}

\subsection{Generalization Experiments}

Secondly, we evaluate the generalization ability of our proposed method emphatically, where we mainly evaluate models according to the ability to detect fake images generated by methods not included in the training phase. We implement experiments on GGFI and FF++ respectively. There are four kinds of manipulation methods both in GGFI and FF++. Each time we remove training images generated by one of four methods in the training phase. For instance, as demonstrated in Table.\ref{tab:general_datasets}, we remove SGAN2 from GGFI, so the trained model will not learn any manipulated images generated by SGAN2. Then, we evaluate the model on the unseen test set containing fake images generated by SGAN2. Table.\ref{tab:unseen_test_GGFI} and Table.\ref{tab:unseen_test_FF++} summarize the experiment results. It is clear to see that our approach significantly outperforms other state-of-the-art methods. Especially when removing PGGAN or SGAN2 from GGFI and removing FS from FF++ in training sets, the test accuracy of our model is much higher than contrasting methods. The results verify that our model is more generalized than others.

\begin{table}[h]
\setlength\tabcolsep{1pt}
  \caption{An example dataset for generalization experiments: Unseen Test Set is used to evaluate methods in generalization experiments. We also use this example dataset in ablation study and evaluate the method both on Unseen Test Set and Test Set.}
  \label{tab:general_datasets}
  \centering
  \ninept
  \begin{tabular}{c|cc|cccc} 
  \hline
     & \multicolumn{2}{c|}{Real} & \multicolumn{4}{c}{Manipulated}\\ 
    & C-HQ & FFHQ & PGGAN & SGAN& SGAN2 & MSGGAN\\
    \hline
    Training Set  & 7500  & 7500 & 5000  & 5000 & 0 & 5000  \\
    Validation Set   & 750  & 750 & 500  & 500 & 0 & 500  \\
    Unseen Test Set  & 2500  & 2500 & 0  & 0 & 5000 & 0  \\
    \cdashline{1-7}[0.8pt/2pt]
    Test Set  & 7500  & 7500 & 5000  & 5000 & 0 & 5000  \\
    \hline
  \end{tabular}
\end{table}

\begin{table}[h]
\setlength\tabcolsep{1.5pt}
  \caption{Generalization Experiments on GGFI.}
  \label{tab:unseen_test_GGFI}
  \centering
  \ninept
  \begin{tabular}{l|p{1.1cm}<{\centering}|p{1cm}<{\centering}|p{1.1cm}<{\centering}|p{1.4cm}<{\centering}} 
  \hline
    Method $\backslash$ Unseen Method  & PGGAN & SGAN & SGAN2 & MSGGAN \\
    \hline
     VGG-19\cite{simonyan2014very} & 80.68 & 98.04 & 60.94 & 89.72  \\
     Xception\cite{Chollet_2017_CVPR} & 87.98 & 97.22 & 59.00 & 93.71  \\
     Detect-VGG-19\cite{Dang_2020_CVPR} & 50.05 & 97.36 & 65.27 &73.43  \\
     Detect-Xception\cite{Dang_2020_CVPR} & 51.22 & 97.33 & 72.05 &83.87  \\
     Capsule\cite{nguyen2019use} & 85.26 & 98.23 & 65.34 & 91.67  \\
     Multi\cite{multi-task}  & 49.84 & 62.38 & 50.94 & 64.35\\
     Patch\cite{chai2020makes} & 50.01 & 93.89 & 79.62 & 94.32  \\
     \textbf{Ours} &  \textbf{93.34} & \textbf{99.60} & \textbf{93.50}  & \textbf{97.18}  \\
    \hline
  \end{tabular}
\end{table}

\begin{table}[t]
\setlength\tabcolsep{2pt}
  \caption{Generalization Experiments on FF++.}
  \label{tab:unseen_test_FF++}
  \centering
  \ninept
 
  \begin{tabular}{l|p{1cm}<{\centering}|p{1cm}<{\centering}|p{1cm}<{\centering}|p{1cm}<{\centering}}
   \hline
    Method $\backslash$ Unseen Method & FS & DF & F2F & NT \\
    \hline
     VGG-19\cite{simonyan2014very} & 49.95  &98.46   &  95.13  &  85.58 \\
     Xception\cite{Chollet_2017_CVPR}&  51.25  & 98.99  & 93.78  & 85.01  \\
     Detect-VGG-19\cite{Dang_2020_CVPR} & 53.79  & 97.95 & 97.93  & 88.68 \\
     Detect-Xception\cite{Dang_2020_CVPR}  & 52.31  & 98.86  &  96.97  & 81.78  \\
     Capsule\cite{nguyen2019use} & 49.88  & 89.69  & 91.53  & 86.96  \\
     Multi\cite{multi-task}  & 85.61 & 98.73  & 98.26  & 93.64\\
     Patch\cite{chai2020makes} & 81.51 & 91.26 & 97.73 & 97.57  \\
     \textbf{Ours} & \textbf{94.47}  & \textbf{99.95} &  \textbf{99.94}  &  \textbf{99.84} \\
     \hline
  \end{tabular}
\end{table}


\subsection{Ablation Study}
Finally, we evaluate the effectiveness of each part of our model. There are three key parts in our proposed method, which are MFSS, LAMs, and SAM. We apply ablation study to them respectively. The dataset for the ablation study is shown in Table.\ref{tab:general_datasets}, and we both test the method on Test Set and Unseen Test Set. 

\textbf{MFSS: }In ablation study, each time we remove one of six fragments from the method to verify the effectiveness of each fragment. Results in Table.\ref{tab:FSS} demonstrate that all six fragments contribute to the final prediction, especially for the generalization ability. By and large, local semantic fragments contribute more than global fragments, which means local fragments contain more discriminative features. 

\begin{table}[!h]
  \caption{Ablation study for MFSS.}
  \label{tab:FSS}
  \centering
  \ninept
  \begin{tabular}{c|cc}
  \hline
    Removed Fragment & Test Set & Unseen Test Set \\ 
    \hline
    Background  & 99.92  & 88.41 \\
    Face & 99.91  & 91.53\\
    Eyes & 99.90  & 90.41\\
    Nose & 99.88  & 87.89\\
    Mouth & 99.79  & 84.88\\
    Pic & 99.91  & 89.54\\
    \cdashline{1-3}[0.8pt/2pt]
    \textbf{None (Our method)} & \textbf{99.95}  & \textbf{93.50}\\
    \hline
  \end{tabular}
\end{table}

\textbf{LAMs and SAM: }Table.\ref{tab:LAM and SAM} summarizes the performance of our approach with and without two proposed attention modules. The results demonstrate that LAMs effectively improve the generalization ability by highlighting discriminative regions, and SAM also improves the performance of our approach. Only with both LAMs and SAM can the method reach the best performance.

\begin{table}[!h]
  \caption{Ablation study for LAMs and SAM.}
  \label{tab:LAM and SAM}
  \centering
  \ninept
  \begin{tabular}{l|cc}
  \hline
    Attention Modules & Test Set & Unseen Test Set \\ 
    \hline
    None  & 99.88  & 76.90 \\
    SAM  & 99.89 & 80.91\\
    LAMs  & 99.91 & 91.26\\
    \textbf{LAMs \& SAM}  & \textbf{99.95}  & \textbf{93.50}\\
   \hline
  \end{tabular}
\end{table}

\section{Conclusion}
This paper tackles the generalization problem of manipulated face image detection. We propose a novel detection method with Multilevel Facial Semantic Segmentation and Cascade Attention Mechanism. MFSS generates six fragments of face images to provide local and global semantic features for detection. Cascade Attention Mechanism consists of LAMs and SAM. LAMs highlight generalized discriminative regions in semantic fragments and SAM measures the contribution of each fragment to the final prediction. To evaluate our approach, we reconstruct two datasets and collect two open-source datasets. Experiments on four datasets demonstrate that our method outperforms state-of-the-arts on various manipulation methods. Moreover, the generalization ability of ours is significantly higher than other methods.


\section{Reference}
\bibliographystyle{IEEEbib}

\end{document}